%% file: main.tex
\documentclass[10pt,twocolumn,letterpaper]{article}

\usepackage[pagenumbers]{cvpr} 
\usepackage[ruled]{algorithm2e}
\usepackage{pifont}
\usepackage{multirow}

\newtheorem{remark}{Remark}[section]
\input{preamble}

\usepackage[normalem]{ulem}
\useunder{\uline}{\ul}{}

\usepackage[marginal]{footmisc}

\definecolor{cvprblue}{rgb}{0.21,0.49,0.74}
\usepackage[pagebackref,breaklinks,colorlinks,citecolor=cvprblue]{hyperref}
\usepackage{multirow}
\newcommand{\tabincell}[2]{\begin{tabular}{@{}#1@{}}#2\end{tabular}}
\def\paperID{15255} 
\def\confName{CVPR}
\def\confYear{2024}

\title{HIR-Diff: Unsupervised Hyperspectral Image Restoration Via Improved Diffusion Models}

\author{Li Pang$^{1,3,}$\footnotemark[1], Xiangyu Rui$^{2,3,}$\thanks{Equal contribution}, Long Cui$^{1}$,Hongzhong Wang$^{1}$, Deyu Meng$^{2,3}$, Xiangyong Cao$^{1,3,}$\thanks{Corresponding author (caoxiangyong@xjtu.edu.cn)}\\
$^{1}$ School of Computer Science and Technology, Xi’an Jiaotong University, Xi'an, China\\
$^{2}$  School of Mathematics and Statistics, Xi’an Jiaotong University, Xi'an, China\\
$^{3}$ Ministry of Education Key Laboratory for Intelligent Networks and Network Security, \\Xi’an Jiaotong University, Xi'an, China\\
}

\begin{document}
\maketitle

\input{sec/0_abstract}
\input{sec/1_intro}

\input{sec/2_background}
\input{sec/4_methodology}

\input{sec/5_experiment}

\input{sec/6_conclusion}

{
    \small
    \bibliographystyle{ieeenat_fullname}
    \bibliography{main}
}

\input{sec/X_supplementary}

\end{document}

%% file: preamble.tex
\usepackage[dvipsnames]{xcolor}


%% file: sec/0_abstract.tex
\begin{abstract}
Hyperspectral image (HSI) restoration aims at recovering clean images from degraded observations and plays a vital role in downstream tasks. Existing model-based methods have limitations in accurately modeling the complex image characteristics with handcraft priors, and deep learning-based methods suffer from poor generalization ability. To alleviate these issues, this paper proposes an unsupervised HSI restoration framework with pre-trained diffusion model (HIR-Diff), which restores the clean HSIs from the product of two low-rank components, i.e., the reduced image and the coefficient matrix. Specifically, the reduced image, which has a low spectral dimension, lies in the image field and can be inferred from our improved diffusion model where a new guidance function with total variation (TV) prior is designed to ensure that the reduced image can be well sampled. The coefficient matrix can be effectively pre-estimated based on singular value decomposition (SVD) and rank-revealing QR (RRQR) factorization. Furthermore, a novel exponential noise schedule is proposed to accelerate the restoration process (about 5$\times$ acceleration for denoising) with little performance decrease. Extensive experimental results validate the superiority of our method in both performance and speed on a variety of HSI restoration tasks, including HSI denoising, noisy HSI super-resolution, and noisy HSI inpainting. The code is available at \href{https://github.com/LiPang/HIRDiff}{https://github.com/LiPang/HIRDiff}. 
\end{abstract} 

%% file: sec/1_intro.tex
\section{Introduction}
\label{sec:introduction}

Hyperspectral imaging is an advanced imaging technique that captures and processes a large number of narrow, contiguous spectral bands across the electromagnetic spectrum. Thus Hyperspectral images (HSIs) can provide more abundant spatial and spectral information than natural images and have extensive applications in various fields, e.g., agriculture~\cite{dale2013hyperspectral, khanal2020remote}, mineral exploration~\cite{shirmard2022review, liu2013targeting} and environmental monitoring~\cite{li2020review}. However, due to sensor limitations and atmospheric interference, HSIs suffer various degeneration during the acquisition process, significantly impairing the performance of downstream tasks. Therefore, HSI restoration is very important for HSI applications and numerous methods have emerged in the past decades.

Existing HSI restoration methods can be mainly categorized into two classes, i.e. model-based approaches~\cite{zhang2013hyperspectral, xiong2019hyperspectral, yuan2019tensor} and deep learning (DL)-based approaches~\cite{bodrito2021trainable, rui2021learning, gong2022learning, pang2022trq3dnet}. Model-based methods reformulate the image inverse problem as an optimization problem which contains data fidelity term and regularization term. The data fidelity term ensures that the restored image is close to the observed image and the regularization term constrains the solving space by exploiting the prior knowledge of the clean images, e.g. low-rank and total variation property. This type of method obtains good generalization ability but the handcraft priors are always subjective and thus cannot fully reveal image characteristics, leaving a large room for improvement. Additionally, the optimization problem could be very complex, resulting in high time costs and suboptimal solutions.

In the past decade, extensive DL-based approaches have been proposed for HSI restoration and these methods can learn the image structure and details from a substantial amount of clean and degraded image pairs. Despite the promising performance, the DL-based methods suffer from a poor generalization problem. Besides, training deep neural networks (DNNs) requires a lot of image pairs, but HSIs are very precious and only limited data are available in most cases. Recently, extensive diffusion-based models have emerged for image restoration tasks such as denoising~\cite{kawar2022denoising}, super-resolution~\cite{choi2021ilvr, kawar2022denoising, saharia2022image, wang2022zero} and inpainting~\cite{kawar2022denoising, lugmayr2022repaint, saharia2022palette, song2020score, wang2022zero}. Among these methods, DDRM~\cite{kawar2022denoising} solves linear inverse problems by performing singular value decomposition (SVD) of the linear degradation matrix and performing diffusion in the spectral space with a pre-trained diffusion model. However, the method cannot be applied directly for HSI restoration since HSIs have more bands than natural images, and the number of bands varies in different datasets due to different sensors. After that, DDS2M~\cite{miao2023dds2m} presents a self-supervised diffusion model for HSI restoration, which restores HSI only using the degraded HSI and can be adaptive to different HSI datasets. Although demonstrating desirable HSI restoration performance and superior generalization ability, DDS2M fails to utilize prior knowledge of available datasets and takes a lot of time to complete the self-supervised training process for each dataset.

To alleviate these issues, inspired by~\cite{rui2023unsupervised} which employs pre-trained diffusion models for pansharpening, we propose an unsupervised HSI restoration framework with pre-trained diffusion model (HIR-Diff). We start from the point that HSIs can be restored from the product of two low-rank components, i.e. the reduced image and the coefficient matrix, which can be estimated separately. Concretely, the reduced image which has a low spectral dimension is defined as several linearly independent bands from the HSI and can be restored using our improved diffusion model, where we propose a new guidance function and a novel exponential noise schedule. The guidance function consisting of a data fidelity term and a total variation (TV) regularization ensures that the reduced image can be better estimated during the diffusion sampling process. The exponential noise schedule can accelerate the diffusion process, that is, it can enable the reduced image to be restored within 20 sampling steps (5$\times$ acceleration for denoising) with little performance decrease. The coefficient matrix, which is needed for the restoration of the reduced image, can be pre-estimated from the degraded image and a predefined band index of the reduced image without any additional information. In this work, we resort to singular value decomposition (SVD) to estimate the coefficient matrix since SVD is noise-robust. Besides, the predefined band index of the reduced image needs to be carefully designed so that each band in the reduced image contains different contents and the estimated coefficient matrix is robust to perturbations. To this end, the ranking-revealing QR (RRQR) factorization~\cite{gu1996efficient} which can identify the numerical rank of a matrix, is adopted to determine the band selection index. 

In summary, our contributions are as follows.
\begin{itemize}
\setlength{\leftskip}{1em}
\item{We propose an unsupervised HSI restoration framework with an improved diffusion model (HIR-Diff), which recovers the clean HSIs from the product of the reduced image and the coefficient matrix. Additionally, a new guidance function with TV prior is designed in the reverse sampling process of the diffusion model to ensure the reduced image can be well sampled.}
\item{We propose an efficient and noise-robust method to estimate the coefficient matrix utilizing SVD and RRQR factorization. Specifically, the RRQR is adopted to predefine the band index of the reduced image so that the estimated coefficient matrix is robust to perturbations.}
\item{We propose a novel exponential noise schedule, which can significantly accelerate the diffusion process compared with existing noise schedules (e.g., 5$\times$ acceleration for denoising) with little performance decrease.}
\end{itemize}

%% file: sec/2_background.tex
\section{Preliminaries}
\label{sec:background}

\subsection{Diffusion Models}
\label{sec:background_diffusion}
Diffusion models are probabilistic generative models that capture the underlying dynamics of data evolution over time~\cite{ho2020denoising, song2019generative}. Owing to the powerful generation ability, diffusion models have gained considerable attention in various tasks, e.g., audio and text generation~\cite{huang2022fastdiff, kong2020diffwave}. A typical diffusion process contains a $T$-step forward process and a $T$-step reverse process. The forward process starts with a clean image and progressively adds noise over $T$ steps, while the reverse process operates in the opposite direction. Given a noise scheduler $\{\bar{\alpha}_t\}_{t=1}^T$ and a clean data $x_0$, adding Gaussian noise to $x_0$ iteratively over $t$ iterations yields
\begin{equation}\label{eq:zt}
\mathbf{x}_t=\sqrt{\bar{\alpha}_t}\mathbf{x}_0+(\sqrt{1-\bar{\alpha}_t})\epsilon,\epsilon\sim\mathcal{N}(0,\mathbf{I}).
\end{equation}
A diffusion model $\epsilon_{\theta}$ is trained to predict the noise as
\begin{equation}\label{eq:epsilon}
\epsilon_\theta(\mathbf{x}_t,t)\approx\epsilon=\frac{\mathbf{x}_t-\sqrt{\bar{\alpha}_t}\mathbf{x}_0}{\sqrt{1-\bar{\alpha}_t}}.
\end{equation}
The reverse process starts with random noise and progressively refines the sample over $T$ steps. 
Recently numerous approaches are proposed to accelerate the sampling process of diffusion models.
A typical method is the denoising diffusion implicit model (DDIM)~\cite{song2020denoising}, where the forward process is modeled as non-Markovian. The model predicts the starting point as
\begin{equation}\label{eq:z0}
\hat{\mathbf{x}}_0=\frac{\mathbf{x}_t-(\sqrt{1-\bar{\alpha}_t})\epsilon_\theta(\mathbf{x}_t,t)}{\sqrt{\bar{\alpha}_t}},
\end{equation}
and then $\mathbf{x}_{t-1}$ is sampled as
\begin{equation}\label{eq:zt_1}
\mathbf{x}_{t-1}=\sqrt{\bar{\alpha}_{t-1}}\hat{\mathbf{x}}_0+(\sqrt{1-\bar{\alpha}_{t-1}})\epsilon_\theta.
\end{equation}
We adopt DDIM as the sampling method in this work since this model can accelerate sampling without significant performance degradation.

\begin{figure*}[ht]
  \centering
  \includegraphics[width=0.9\textwidth]{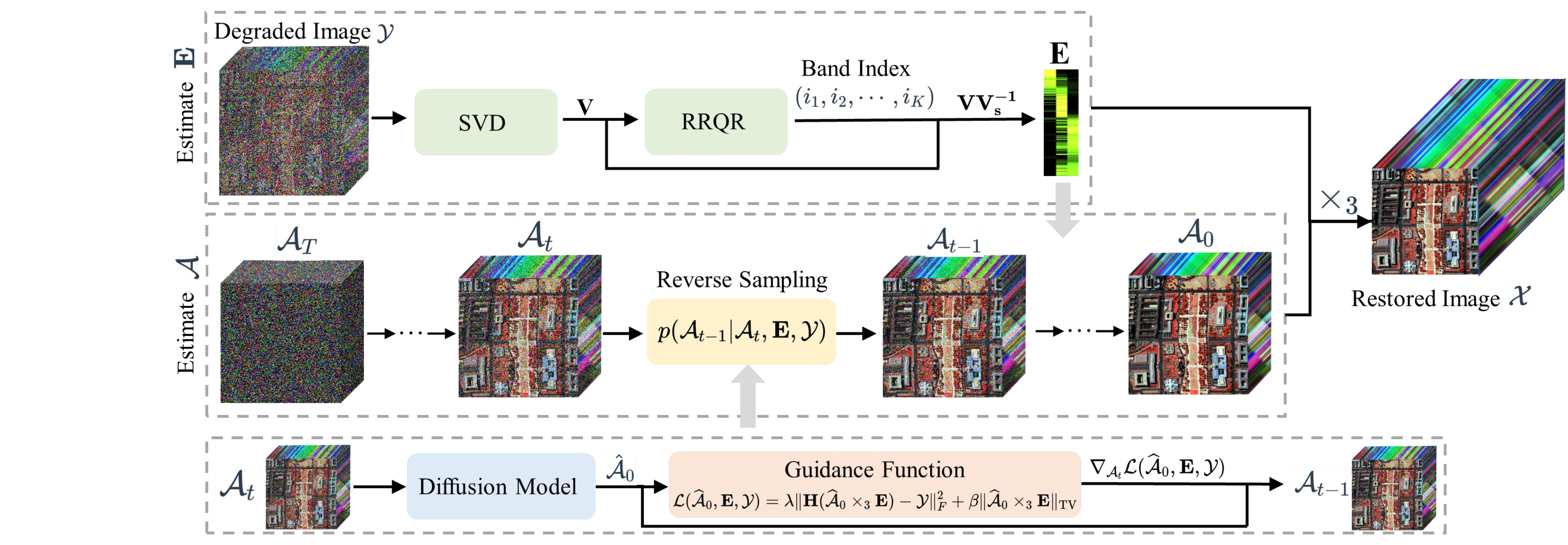}
  \caption{The overall framework of the proposed HIR-Diff. First, the coefficient matrix $\mathbf{E}$ is estimated from the degraded image using SVD and RRQR. Then, taking the degraded image and the estimated matrix $\mathbf{E}$ as conditions, the reduced image $\mathcal{A}$ is reconstructed with an improved pre-trained diffusion model that contains a newly designed guidance function. Finally, the clean image is restored from the product of the estimated  $\mathbf{E}$ and $\mathcal{A}$.}\label{fig:diffusion}
\end{figure*}

\subsection{Hyperspectral Image Restoration}
\label{sec:background_hir}
HSI restoration task aims to recover clean images from degraded observations and its degradation process is
\begin{equation}\label{eq:restoration}
\mathcal{Y} = \mathbf{H}(\mathcal{X}) + \mathcal{Z},
\end{equation}
where $\mathcal{Y}$ is the degraded image, $\mathcal{X}$ is the clean HSI, $\mathbf{H}$ is a known degradation operation, and $\mathcal{Z}$ denotes an additive Gaussian noise. For the denoising task, $\mathbf{H}$ is an identity operation. For the super-resolution task, $\mathbf{H}$ contains a blur operation and a down-sampling operation. For the inpainting task, $\mathbf{H}$ denotes a random mask.


%% file: sec/4_methodology.tex
\section{Proposed Model}
\label{sec:methodology}

\subsection{Notations}
The scalar, vector, matrix and tensor are defined as $x, \mathbf{x}, \mathbf{X}$ and $\mathcal{X}$, respectively. Given a third-order tensor $\mathcal{X}\in\mathbb{R}^{H\times W\times K}$ and a matrix $\mathbf{E}\in\mathbb{R}^{B\times K}$, mode-3 multiplication can be defined as $\mathcal{Y}=\mathcal{X}\times_3\mathbf{E}$, resulting in a new tensor $\mathcal{Y}\in\mathbb{R}^{H\times W\times B}$.
$\mathbf{X}_{(3)} \in \mathbb{R}^{B\times HW}$ denotes the mode-3 unfolding of $\mathcal{X}$ and is obtained by flattening the first two dimensions of $\mathcal{X}$. And $\mathbf{fold}_3(\mathbf{X}_{(3)})$ means reshaping $\mathbf{X}_{(3)}$ back to $\mathcal{X}$.

\subsection{Overall Framework}
\label{subsec:overall_framework}
Due to strong spectral correlation, HSIs thus exhibit low-rank property, which has been widely used for HSI restoration~\cite{letexier2008noise, he2019non, cao2016robust}. Following this research line, we assume that the clean HSI $\mathcal{X} \in \mathbb{R}^{H\times W\times B}$ can be recovered as
\begin{equation}\label{eq:lowrank}
\mathcal{X} = \mathcal{A} \times_3 \mathbf{E},
\end{equation}
where $\mathcal{A} \in \mathbb{R}^{H\times W\times K}$ denotes the reduced image, $\mathbf{E} \in \mathbb{R}^{B \times K}$ denotes the coefficient matrix and $K \ll B$. Therefore, to restore the clean HSI, we first need to estimate the reduced image $\mathcal{A}$ and the coefficient matrix $\mathbf{E}$. In this work, we propose a method to separately estimate $\mathcal{A}$ and $\mathbf{E}$. The overall framework is illustrated in Fig.~\ref{fig:diffusion}. 

Overall, given a degraded HSI $\mathcal{Y}$, we first estimate the coefficient matrix $\mathbf{E}$, which is a necessary condition for the restoration of the reduced image $\mathcal{A}$. Then, we estimate the reduced image $\mathcal{A}$ using our improved diffusion model. $\mathcal{A}$ is supposed to be a sequence of bands selected from the HSI so that the distribution is consistent with the diffusion priors. In other words, defining $(i_1,i_2,...,i_K)$ as the band index, $\mathcal{A}$ satisfies $\mathbf{A}_{(3)} = \left[ \mathbf{x}_{i_1}^\mathrm{T}, \mathbf{x}_{i_2}^\mathrm{T}, \cdots, \mathbf{x}_{i_K}^\mathrm{T} \right]^\mathrm{T}$, where $\mathbf{x}_{i_n}$ denotes the $i_n$th row of $\mathbf{X}_{(3)}$. Given a predefined band index $(i_1,i_2,...,i_K)$ of the reduced image and a degraded image $\mathcal{Y}$, the coefficient matrix $\mathbf{E}$ can be estimated by singular value decomposition (SVD) without any additional information. Once $\mathbf{E}$ is well estimated, $\mathcal{A}$ could be inferred by applying our improved diffusion model with our newly designed guidance function. Moreover, an exponential noise schedule is also proposed to accelerate the sampling.


Although a restored image can be obtained for each possible predefined band selection index, the band index needs to be carefully designed to ensure that each band in $\mathcal{A}$ can encode different contents, which helps to generate a more robust estimation of $\mathbf{E}$ and thus obtain better restoration results. To this end, we resort to a classical rank revealing QR (RRQR) decomposition~\cite{gu1996efficient} to determine the band index. Next, we provide a more detailed description for estimating the coefficient matrix $\mathbf{E}$ and the reduced image $\mathcal{A}$.

\subsection{Coefficient Matrix Estimation}
\label{subsec:matrix_estimation}
\subsubsection{Coefficient Matrix Estimation Using SVD}
As discussed before, the coefficient matrix $\mathbf{E}$ needs to be pre-estimated from an observed degraded image $\mathcal{Y}$ and a predefined band index $(i_1,i_2,...,i_K)$ without any additional information about the reduced image $\mathcal{A}$ and the clean image $\mathcal{X}$. To this end, inspired by~\cite{he2019non}, we resort to SVD to estimate the coefficient matrix $\mathbf{E}$. First, the matrix version of the HSI $\mathcal{Y}$ is decomposed using the rank-$K$ SVD as
\begin{equation}\label{eq:svd}
\mathbf{Y}_{(3)}^\mathrm{T}=\mathbf{(US)V}^\mathrm{T},
\end{equation}
where $\mathbf{U} \in \mathbb{R}^{HW \times K}$, $\mathbf{S} \in \mathbb{R}^{K \times K}$ and $\mathbf{V} \in \mathbb{R}^{B \times K}$. We define $\mathbf{V}_s = \left[ \mathbf{v}_{i_1}^\mathrm{T}, \mathbf{v}_{i_2}^\mathrm{T}, \cdots, \mathbf{v}_{i_K}^\mathrm{T} \right]^\mathrm{T}$, where $\mathbf{v}_{i_n}$ denotes the $i_n$th row of $\mathbf{V}$. Then we can obtain
\begin{equation}\label{eq:rgb}
\mathbf{A}_{\mathbf{Y}(3)}^\mathrm{T} = \mathbf{(US)V_s}^\mathrm{T},
\end{equation}
Combining Eq.~(\ref{eq:svd}) and Eq.~(\ref{eq:rgb}), we have
\begin{align}
\mathbf{Y}_{(3)}^\mathrm{T} &= \mathbf{A}_{\mathbf{Y}(3)}^\mathrm{T}(\mathbf{V_s}^\mathrm{T})^{-1}\mathbf{V}^\mathrm{T}, \label{eq:rgb2hsi_0}
\end{align}
which is equivalent to
\begin{align}
\mathcal{Y} &= \mathcal{A}_\mathbf{Y} \times_3 (\mathbf{VV}_s^{-1}). \label{eq:rgb2hsi}
\end{align}
Then we have the following remark

\begin{remark}\label{remark:E}
Let $\mathbf{Y}_{(3)}^\mathrm{T}=\mathbf{(US)V}^\mathrm{T}$ be the rank-$K$ SVD of the observed HSI $\mathcal{Y}$, and define $\mathbf{V}_s$ as the $(i_1,i_2,...,i_K)$th rows of $\mathbf{V}$ . Then $\mathbf{VV}_s^{-1}$ is equivalent to the expected $\mathbf{E}$ in Eq.~(\ref{eq:lowrank}) when rank $K$ is small.
\end{remark}
Here we provide a rough proof for this remark. Defining $\mathcal{\bar{Y}} = \mathbf{H}(\mathcal{X})$, the degraded image $\mathcal{Y}$ could be regarded as $\mathcal{\bar{Y}}$ with \emph{i.i.d.} Gaussian noise. $\mathcal{\bar{Y}}$ can be decomposed similarly to Eq.~(\ref{eq:rgb2hsi}), namely
\begin{equation}\label{eq:Abar0}
\mathcal{\bar{Y}} = \mathcal{\bar{A}}_\mathbf{Y} \times_3 (\mathbf{\bar{V}\bar{V}}_s^{-1}), 
\end{equation}
where $\mathcal{\bar{A}}_\mathbf{Y}$ is the $(i_1,i_2,...,i_K)$th bands of $\mathcal{\bar{Y}}$, $\mathbf{\bar{V}}$ is the right singular vectors obtained from the SVD of $\mathcal{\bar{Y}}$ and $\bar{\mathbf{V}}_s$ is the $(i_1,i_2,...,i_K)$th rows of $\bar{\mathbf{V}}$. Due to the favourable mathematical property of the SVD, $\mathbf{V}$ is an orthonormal matrix, i.e. $\mathbf{V}^\mathrm{T}\mathbf{V}=\mathbf{I}$. The orthogonal property encourages the representations of $\mathbf{V}$ to be more distinguishable from each other, helping to keep the noise of $\mathbf{V}$~\cite{he2019non}. Additionally, when $K$ is small, there is no significant difference between $\mathbf{V}$ and $\mathbf{\bar{V}}$ as the columns of $\mathbf{V}$ indicate the directions that capture the most significant variations in the data. Therefore, the decomposition of $\mathcal{\bar{Y}}$ can be rewritten as
\begin{equation}\label{eq:Abar}
\mathcal{\bar{Y}} = \mathcal{\bar{A}}_\mathbf{Y} \times_3 (\mathbf{VV}_s^{-1}),
\end{equation}
As the degradation operation $\mathbf{H}$ is linear and is performed in the spatial dimension, then we have
\begin{align}
\mathcal{\bar{Y}} = \mathbf{H}(\mathcal{X}) = \mathbf{H}(\mathcal{A} \times_3 \mathbf{E})= \mathbf{H}(\mathcal{A}) \times_3 \mathbf{E}= \mathcal{\bar{A}}_\mathbf{Y} \times_3 \mathbf{E}. \label{eq:X}
\end{align}
By comparing Eq.~(\ref{eq:Abar}) and Eq.~(\ref{eq:X}), it can be easily seen that the coefficient matrix can be defined as
\begin{equation}\label{eq:E}
\mathbf{E} = \mathbf{VV}_s^{-1}.
\end{equation}

\subsubsection{Band Index Selection Using RRQR}
While the coefficient matrix could be estimated from Eq.~(\ref{eq:E}), the predefined index $(i_1,i_2,...,i_K)$ of the reduced image needs to be carefully selected so that $|\det(\mathbf{V}_s)|>0$ and $\mathbf{V}_s^{-1}$ exists. Actually, $|\det(\mathbf{V}_s)|$ is expected to be large owing to the following two reasons: A larger $|\det(\mathbf{V}_s)|$ indicates that each band in the reduced image $\mathcal{A}$ encodes different image information, improving the HSI restoration performance. In addition, if $|\det(\mathbf{V}_s)|$ is small, the value in the coefficient matrix $\mathbf{E}$ could be very large, which could lead to numerical instability during the estimation process of the reduced image $\mathcal{A}$. To this end, a rank-revealing QR (RRQR) algorithm proposed in~\cite{gu1996efficient} is employed to determine the band selection index. Using the RRQR factorization on the matrix $\mathbf{V}^\mathrm{T}$ can be decomposed as
\begin{equation}\label{eq:rrqrV}
\setlength{\arraycolsep}{1.5pt}
\mathbf{V^\mathrm{T}\mathbf{\Pi}} = \mathbf{QR} \equiv \left[\begin{array}{cc}\mathbf{Q}\mathbf{R}_{1}&\mathbf{Q}\mathbf{R}_{2}\end{array}\right],
\end{equation}
where $\mathbf{\Pi} \in \mathbb{R}^{B\times B}$ is a permutation matrix, $\mathbf{Q} \in \mathbb{R}^{K\times K}$ is an orthogonal matrix and $\mathbf{R} \in \mathbb{R}^{K\times B}$ is an upper triangular matrix. The RRQR factorization algorithm works by interchanging any pair of columns that increases sufficiently $|\det(\mathbf{R_{1}})|$. We define $\mathbf{V}_s^\mathrm{T}$ as $\mathbf{Q}\mathbf{R}_1$ and then we could obtain $|\det(\mathbf{V}_s)|=|\det(\mathbf{R}_{1})|$. Therefore, when the algorithm terminates, $|\det(\mathbf{V}_s)|$ is maximized and the indices corresponding to the first $K$ columns of the permuted matrix $\mathbf{V}^\mathrm{T}\mathbf{\Pi}$ can be defined as the band selection index.
The choice of selection index could not cause $|\det(\mathbf{V}_s)|$ to be too large, which could also result in numerical instability. Concretely, from Hadamard's inequality~\cite{garling2007inequalities}, we have
\begin{equation}\label{eq:hadamard}
|\det(\mathbf{V}_s)|\leq\prod_{i=1}^K\|\mathbf{vs}_i\|_2,
\end{equation}
where $\mathbf{vs}_i$ is the $i$th column of $\mathbf{V}_s$. Since each column $\mathbf{v}$ in matrix $\mathbf{V}$ satisfies $\|\mathbf{v}\|_2=1$, we have $\|\mathbf{vs}_i\|_2 \leq \|\mathbf{v}_i\|_2=1$ and $|\det(\mathbf{V}_s)| \leq 1$.

\subsection{Conditional Denoising Diffusion Model}
\label{subsec:conditional_diffusion}
With the pre-estimated coefficient matrix $\mathbf{E}$ and a predefined band index, we employ a pre-trained diffusion model to estimate the reduced image $\mathcal{A}$. Recently, emerging approaches~\cite{chung2022diffusion, fei2023generative, bansal2023universal} are proposed for image generation with various conditions utilizing a pre-trained diffusion model. In general, given a condition or a guidance $\mathbf{y}$, these approaches model $p_{\theta}(\mathbf{x}_{t-1}|\mathbf{x}_{t},\mathbf{y})$ during the reverse sampling process by introducing gradient on $\epsilon_\theta$ as
\begin{equation}\label{eq:epsilon}
\hat{\epsilon}_\theta(\mathbf{x}_t,t)=\epsilon_\theta(\mathbf{x}_t,t)+s(t)\cdot\nabla_{\mathbf{x}_t}\mathcal{L}(\hat{\mathbf{x}}_0, \mathbf{y}),
\end{equation}
where $s(t)$ denotes the guidance strength in the $t$th step and $\mathcal{L}$ denotes a loss function which measures the distance between the current predicted $\hat{\mathbf{x}}_0$ and the expected image. In our case, taking the degraded image $\mathcal{Y}$ and the pre-estimated coefficient matrix $\mathbf{E}$ as conditions, we employ diffusion models to estimate the reduced image $\mathcal{A}$ as shown in Fig.~\ref{fig:diffusion}. Specifically, Eq.~(\ref{eq:epsilon}) can be rewritten as
\begin{equation}\label{eq:epsilon1}
\hat{\epsilon}_\theta(\mathcal{A}_t,t)=\epsilon_\theta(\mathcal{A}_t,t)+s(t)\cdot\nabla_{\mathcal{A}_t}\mathcal{L}(\hat{\mathcal{A}}_0, \mathbf{E}, \mathcal{Y}),
\end{equation}
where $\hat{\mathcal{A}}_0$ is estimated by Eq.(\ref{eq:z0}) as
\begin{equation}\label{eq:A0}
\hat{\mathcal{A}}_0 = \frac{\mathcal{A}_t-(\sqrt{1-\bar{\alpha}_t})\epsilon_\theta(\mathcal{A}_t,t)}{\sqrt{\bar{\alpha}_t}}.
\end{equation}
Then $\mathcal{A}_{t-1}$ is sampled from Eq.~(\ref{eq:zt}) as
\begin{equation}\label{eq:At}
\mathcal{A}_{t-1} = \sqrt{\bar{\alpha}_{t-1}}\hat{\mathcal{A}}_0+(\sqrt{1-\bar{\alpha}_{t-1}})\hat{\epsilon}_\theta(\mathcal{A}_t,t).
\end{equation}
The loss function $\mathcal{L}(\hat{\mathcal{A}}_0, \mathbf{E}, \mathcal{Y})$ proposed in our work contains a data fidelity term and a TV regularization. The data fidelity term ensures that the predicted image closely matches the observed data. The TV term that has been widely adopted in HSI restoration~\cite{wang2017hyperspectral, peng2020enhanced, peng2022fast} contributes to denoising and edge preservation. Specifically, the guidance function $\mathcal{L}(\hat{\mathcal{A}}_0, \mathbf{E}, \mathcal{Y})$ is defined as
\begin{equation}\label{eq:LossFunction}
\mathcal{L}(\hat{\mathcal{A}}_0, \mathbf{E}, \mathcal{Y}) = \lambda \|\mathbf{H}(\hat{\mathcal{A}}_0 \times_3 \mathbf{E}) - \mathcal{Y} \|_F^2 + \beta \|\hat{\mathcal{A}}_0 \times_3 \mathbf{E}\|_{\text{TV}},
\end{equation}
where $\lambda$ and $\beta$ are hyperparameters. The proposed method is summarized in Algorithm~\ref{alg:algorithm1}. When diffusion sampling finishes, the restored HSI $\mathcal{X}_{0}$ is represented as $\mathcal{A}_{0} \times_3 \mathbf{E}$.


\begin{algorithm}[t]
	\caption{HIR-Diff Method}
	\label{alg:algorithm1}
    \KwIn{$\mathcal{A}_T$ sampled from $\mathcal{N}(0, I)$,  diffusion model $\epsilon_{\theta}$, noise scales $\{\bar{\alpha}_t\}^{T}_{t=1}$, guidance strength $s(t)$, loss function $\mathcal{L}$, degraded HSI $\mathcal{Y}$, estimated coefficient $\mathbf{E}$}
    \KwOut{Clean HSI $\mathcal{X}_0$}

	\BlankLine
    \For{$t = T, T-1, \cdots 1$}{
        \textbf{step 1:} estimate $\hat{\mathcal{A}}_0$ by Eq.~(\ref{eq:A0}) \\
        \textbf{step 2:} calculate $\mathcal{L}(\hat{\mathcal{A}}_0, \mathbf{E}, \mathcal{Y})$ by Eq.~(\ref{eq:LossFunction}) \\
        \textbf{step 3:} estimate $\hat{\epsilon}_\theta(\mathcal{A}_t,t)$  by Eq.~(\ref{eq:epsilon1}) \\
        \textbf{step 4:} sample $\mathcal{A}_{t-1}$ by Eq.~(\ref{eq:At})
    }
    $\mathcal{X}_0 = \mathcal{A}_0 \times_3 \mathbf{E}$\\
    \Return{$\mathcal{X}_0$}
\end{algorithm}

\subsection{Exponential Noise Schedule}
\label{subsec:diffusion_improvement}


\begin{figure}[h]
	\centering
	\includegraphics[width=1\linewidth]{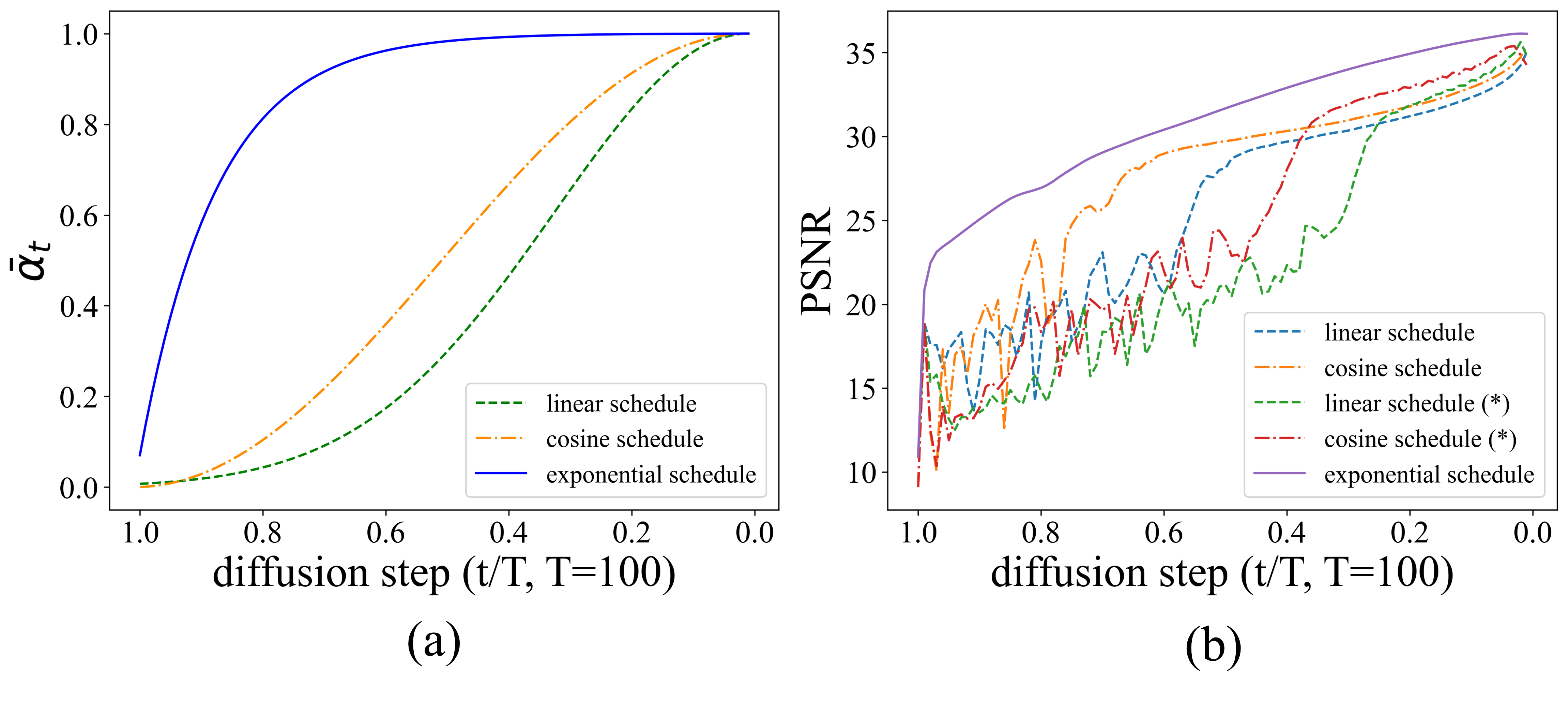}
	\caption{(a) The $\bar{\alpha}_t$ in the linear schedule, cosine schedule and our proposed exponential schedule. (b) The PSNR values throughout the diffusion process with different noise schedules. Linear schedule (*) and cosine schedule (*) denote the results when the guidance strength is enhanced. }
	\label{fig:schedule_psnr}
\end{figure}

We found that when the widely used linear noise schedule~\cite{ho2020denoising} and the cosine noise schedule~\cite{nichol2021improved} are used, the Peak Signal-to-Noise Ratio (PSNR) value of the restored image fluctuates drastically at the beginning and does not converge at the end of the sampling process, as shown in Fig.~\ref{fig:schedule_psnr}. Increasing the sampling steps or improving the strength of the guide does not alleviate the problem. One possible reason is that with the guidance information as illustrated in Eq.~(\ref{eq:epsilon}), the noise rapidly decays at the beginning of the sampling process since the diffusion process starts with random noise while the conditional image contains abundant image details. It can be observed from Eq.~(\ref{eq:zt}) that the noise schedule $\bar{\alpha}_t$ reflects how much information is contained in the current sample. Therefore, the $\bar{\alpha}_t$ is supposed to increase rapidly at the beginning, which is inconsistent with the linear schedule and cosine schedule. Besides, the PSNR value increases rapidly at the end of the sampling process, indicating that a larger $\bar{\alpha}_t$ helps to refine the image details and achieve the local optimum at the end. During this stage, the diffusion prior progressively works to generate a desirable image. Therefore, we design a different noise schedule in terms of $\bar{\alpha}_t$ as
\begin{align}\label{eq:alphas}
\bar{\alpha}_t &= e^{-kt/T}, \quad t=1,2,\cdots,T, \\
\bar{\alpha}_t &= \frac{\bar{\alpha}_t - \min\{\bar{\alpha}_t\}_{t=1}^T}{\max\{\bar{\alpha}_t\}_{t=1}^T - \min\{\bar{\alpha}_t\}_{t=1}^T} \times (1 - \epsilon) + \epsilon, \\
t &= 1,2,\cdots,T.
\end{align}
where $k$ is a hyperparamter and $\epsilon$ is a small value to prevent $\hat{\alpha}_0$ from being zero. Our exponential noise schedule increases rapidly at the beginning and changes slowly at the end so that the random noise can converge to the observed image rapidly and the image details can be well refined.

\textbf{Remark:} It should be noted that our improved diffusion model modifies DDIM in two aspects, i.e. the proposed exponential noise schedule and the designed guidance function introduced in Sec.~\ref{subsec:conditional_diffusion}, and thus providing better sample quality and faster sampling speed. The effectiveness of the two modifications will be discussed in Sec.~\ref{subsec:ablation}.

%% file: sec/5_experiment.tex
\section{Experiment}
\label{sec:experiment}

\subsection{Datasets and Evaluation Metrics}
We evaluate the HSI restoration performance on three publicly available datasets, namely Washington DC (WDC) Mall\footnote{\url{http://lesun.weebly.com/hyperspectral-data-set.html}} whose size is $1208 \times 307 \times 191$, Houston\footnote{\url{https://hyperspectral.ee.uh.edu/?page_id=459}} whose size is $349 \times 1905 \times 144$, and Salinas\footnote{\url{https://www.ehu.eus/ccwintco/index.php/Hyperspectral_Remote_Sensing_Scenes}} whose size is $512 \times 217 \times 224$. For each dataset, we crop the center area and remove some noisy bands, deriving three HSIs with size $256 \times 256 \times 191$, $256 \times 256 \times 124$ and $128 \times 128 \times 190$, respectively. Two commonly used evaluation metrics, i.e. peak signal-to-noise ratio (PSNR) and structure similarity (SSIM), are adopted to evaluate the performance.

\subsection{Competing Methods}
\paragraph{HSI Denoising:} The HSI denoising aims at recovering the clean HSI from its noisy observation. We mainly consider Gaussian noise, and the standard deviation of Gaussian noise $\sigma$ in the range of [0, 255] is set as 30, 50, and 70, respectively. Three model based methods including BM4D~\cite{maggioni2012nonlocal}, NGMeet~\cite{he2019non}, ETPTV~\cite{chen2023hyperspectral}, four deep learning based methods including T3SC~\cite{bodrito2021trainable}, MACNet~\cite{xiong2021mac}, SST~\cite{li2023spatial}, SERT~\cite{li2023spectral} and an unsupervised deep learning based methods (i.e. DDS2M~\cite{miao2023dds2m}) are adopted for comparison. For the DL-based methods, we use available models\footnote{\url{https://github.com/MyuLi/SERT}} trained on ICVL\footnote{\url{https://icvl.cs.bgu.ac.il/hyperspectral/}} with Gaussian noise for comparison.
\paragraph{Noisy HSI Super-Resolution:} The noisy HSI super-resolution aims at recovering the clean HSI from a noisy low-resolution observation. Specifically, the low-resolution image is obtained by first spatially blurring the clean image using a Gaussian-shape filter, then downsampling the blurred image and finally adding the image with Gaussian noise with $\sigma=30$. Seven methods, including two unsupervised deep learning based methods, i.e. DIP2d~\cite{sidorov2019deep} and DIP3d~\cite{sidorov2019deep}, and five supervised deep learning based methods, i.e. SFCSR~\cite{wang2020hyperspectral}, SSPSR~\cite{jiang2020learning}, MCNet~\cite{li2020mixed}, RFSR~\cite{wang2021hyperspectral}, PDENet~\cite{hou2022deep} are adopted for comparison. Since the pre-trained model is not available, we train all the supervised models on CAVE\footnote{\url{https://www.cs.columbia.edu/CAVE/databases/multispectral/}} with the setting proposed in~\cite{wang2020hyperspectral}.

\paragraph{Noisy HSI Inpainting:} The noisy HSI inpainting aims at recovering the clean HSI from noisy and incomplete data. Concretely, we randomly mask a portion of pixels and add the Gaussian noise ($\sigma=30$) to the observed area. The mask rate is set as 0.7, 0.8 and 0.9, respectively. Four model-based methods (i.e. TRPCA~\cite{lu2019tensor}, TRLRF~\cite{yuan2019tensor}, S2NTNN~\cite{luo2022self} and HLRTF~\cite{luo2022hlrtf}) and three unsupervised DL-based methods (i.e. DIP2d~\cite{sidorov2019deep}, DIP3d~\cite{sidorov2019deep} and DDS2M~\cite{miao2023dds2m}) are adopted for comparison.

\subsection{Implementation Details}
A pre-trained diffusion model\footnote{\url{https://github.com/wgcban/ddpm-cd}}
trained on a large amount of 3-channel (i.e. RGB) remote sensing images~\cite{gedara2022remote} is used to generate the reduced image $\mathcal{A}$. The diffusion sampling step $T$ is set as 20 for all the HSI restoration tasks. All DL-based models are trained on an NVIDIA Geforce RTX 3090 GPU.

\subsection{Results for HSI Restoration}
\begin{figure*}[h]
  \centering
  \includegraphics[width=0.8\textwidth]{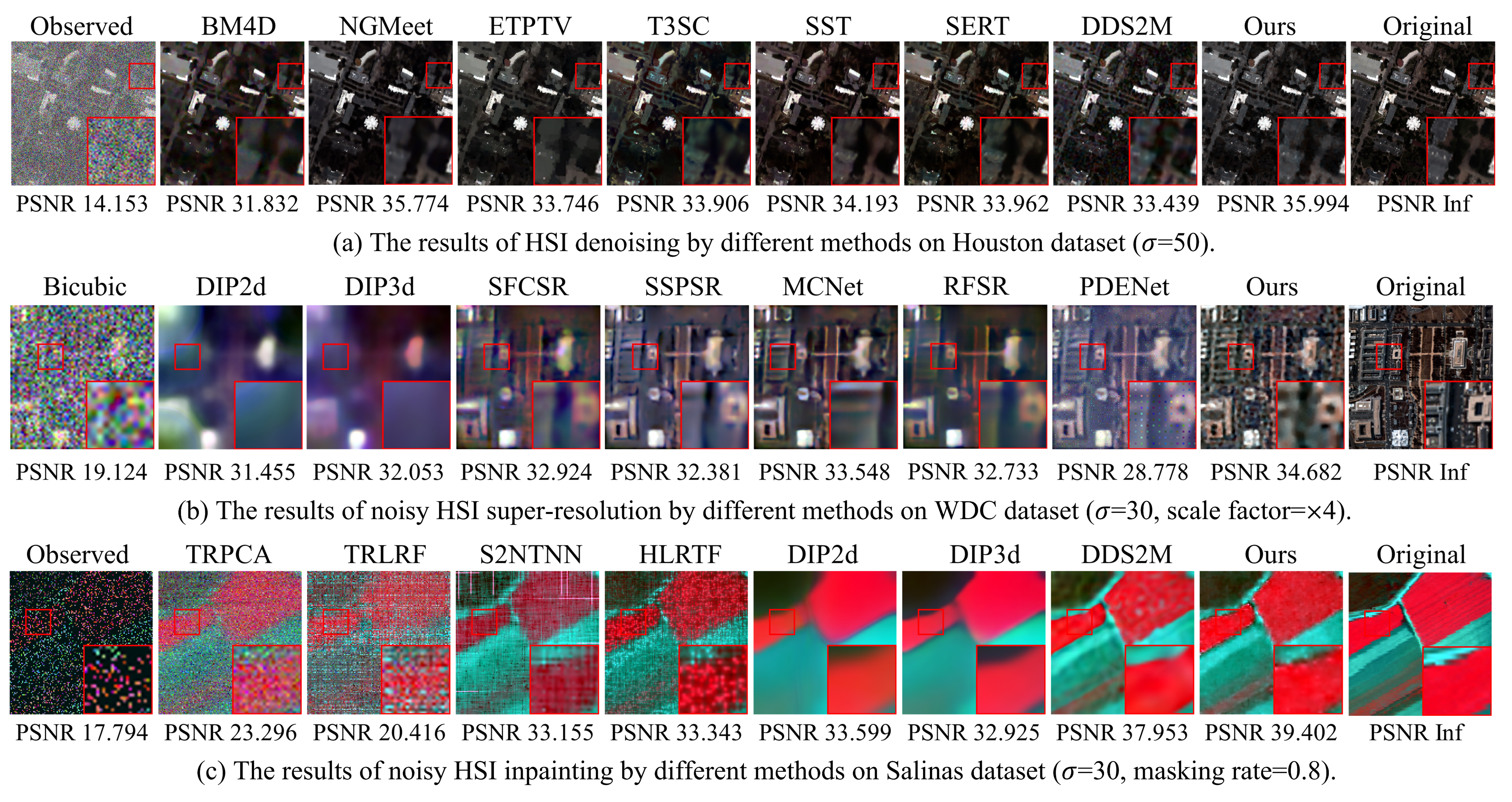}
  \caption{The visual result comparison of all the competing methods on the HSI restoration task.}\label{fig:result}
\end{figure*}

The results of different methods on HSI denoising, noisy HSI super-resolution and noisy HSI inpainting are demonstrated in Table~\ref{table:denoise}, Table~\ref{table:sr} and Table~\ref{table:inpainting}, respectively. It can be seen that our proposed method demonstrates superior HSI restoration performance compared to other competitive models. Compared with model-based approaches (e.g., BM4D and ETPTV) which highly rely on handcraft priors, the pre-trained diffusion model employed in our method is able to capture complex intrinsic characteristics and image details from abundant data. The supervised DL-based methods suffer limited generalization ability and struggle with the restoration task of unseen data. In contrast, our model demonstrates desirable generalization ability as the employed diffusion model learns the image structure and details in a self-supervised manner, which enables the model to be competitive for various tasks and datasets. In addition, prior knowledge including low rank and total variation is introduced in our model, which helps to regularize the restoration process and thus obtain better restoration results. Moreover, other unsupervised DL-based methods, e.g., DIP2d, DIP3d and DDS2M, take lots of time to exploit the image inherent structure with an untrained neural network for each individual dataset, while our method benefitting from self-supervised training is considerably more efficient and fast. Some visual results for different tasks are shown in Fig.~\ref{fig:result}, from which it can be observed that our method can restore more accurate and reliable visual results.

\begin{table}[h]
\caption{The average quantitative results for HSI denoising. The \textbf{best} and \underline{second-best} values are highlighted.}
\label{table:denoise}
\renewcommand\arraystretch{1.2}
\resizebox{\linewidth}{!}{
\begin{tabular}{cl|ccccccc}
\hline
\multicolumn{2}{c|}{standard deviation} & \multicolumn{2}{c}{30}         & \multicolumn{2}{c}{50}         & \multicolumn{2}{c}{70}         & \multirow{2}{*}{Time (s)} \\
\multicolumn{1}{l}{Dataset}   & Method  & PSNR           & SSIM          & PSNR           & SSIM          & PSNR           & SSIM          &                           \\ \hline
\multirow{8}{*}{WDC mall}     & BM4D    & 37.09          & 0.90          & 34.48          & 0.83          & 32.81          & 0.77          & 286                       \\
                              & NGMeet  & \textbf{43.77} & \textbf{0.98} & \textbf{41.07} & \textbf{0.96} & \textbf{39.82} & \textbf{0.94} & 65                        \\
                              & ETPTV   & 41.08          & 0.95          & 37.12          & 0.91          & 35.15          & 0.87          & 629                       \\
                              & T3SC    & 38.70          & 0.92          & 37.04          & 0.88          & 35.94          & 0.86          & 3                         \\
                              & SST     & 39.09          & 0.93          & 37.24          & 0.90          & 36.19          & 0.87          & 5                         \\
                              & SERT    & 38.98          & 0.93          & 37.08          & 0.89          & 35.90          & 0.86          & 5                         \\
                              & DDS2M   & 41.58          & 0.95          & 39.13          & 0.93          & 38.83          & 0.92          & 3132                      \\
                              & Ours    & {\ul 42.85}    & {\ul 0.97}    & {\ul 40.77}    & {\ul 0.94}    & {\ul 39.33}    & {\ul 0.92}    & 17                        \\ \hline
\multirow{8}{*}{Houston}      & BM4D    & 33.89          & 0.84          & 31.83          & 0.77          & 30.57          & 0.72          & 184                       \\
                              & NGMeet  & \textbf{38.50} & \textbf{0.94} & {\ul 35.77}    & {\ul 0.89}    & {\ul 34.36}    & {\ul 0.85}    & 62                        \\
                              & ETPTV   & 35.78          & 0.90          & 33.75          & 0.84          & 32.47          & 0.80          & 322                       \\
                              & T3SC    & 35.74          & 0.90          & 33.91          & 0.85          & 32.75          & 0.81          & 3                         \\
                              & SST     & 36.26          & 0.92          & 34.19          & 0.87          & 32.87          & 0.83          & 6                         \\
                              & SERT    & 36.11          & 0.91          & 33.96          & 0.86          & 32.58          & 0.82          & 4                         \\
                              & DDS2M   & 35.64          & 0.91          & 33.44          & 0.85          & 31.96          & 0.79          & 1813                      \\
                              & Ours    & {\ul 38.13}    & {\ul 0.94}    & \textbf{36.01} & \textbf{0.90} & \textbf{34.56} & \textbf{0.86} & 25                        \\ \hline
\multirow{8}{*}{Salinas}      & BM4D    & 38.62          & 0.91          & 35.96          & 0.86          & 34.23          & 0.82          & 68                        \\
                              & NGMeet  & \textbf{44.76} & \textbf{0.98} & \textbf{42.23} & \textbf{0.96} & \textbf{40.96} & \textbf{0.95} & 16                        \\
                              & ETPTV   & 42.52          & 0.95          & 40.44          & 0.93          & 38.03          & 0.91          & 91                        \\
                              & T3SC    & 41.06          & 0.95          & 39.47          & 0.94          & 38.36          & 0.92          & 3                         \\
                              & SST     & 38.93          & 0.94          & 37.51          & 0.92          & 36.39          & 0.91          & 4                         \\
                              & SERT    & 38.80          & 0.94          & 37.26          & 0.92          & 36.07          & 0.90          & 3                         \\
                              & DDS2M   & 43.29          & 0.96          & 40.05          & 0.93          & 38.10          & 0.90          & 846                       \\
                              & Ours    & {\ul 43.79}    & {\ul 0.96}    & {\ul 41.95}    & {\ul 0.95}    & {\ul 39.48}    & {\ul 0.93}    & 13                        \\ \hline
\end{tabular}
}
\end{table}

\begin{table}[h]
\caption{The average quantitative results for noisy HSI super-resolution. The \textbf{best} and \underline{second-best} values are highlighted.}
\label{table:sr}
\renewcommand\arraystretch{1.2}
\resizebox{\linewidth}{!}{
\begin{tabular}{cl|ccccccc}
\hline
\multicolumn{2}{c|}{Scale}           & \multicolumn{2}{c}{$\times2$}  & \multicolumn{2}{c}{$\times4$}  & \multicolumn{2}{c}{$\times8$}        & \multirow{2}{*}{Time (s)} \\
\multicolumn{1}{l}{Dataset} & Method & PSNR           & SSIM          & PSNR           & SSIM          & PSNR           & SSIM                &                           \\ \hline
\multirow{8}{*}{WDC mall}   & DIP2d  & 32.18          & 0.58          & 31.62          & 0.57          & 29.92          & 0.53                & 206                       \\
                            & DIP3d  & 31.81          & 0.57          & 32.05          & 0.58          & 30.10          & 0.51                & 16644                     \\
                            & SFCSR  & 33.74          & 0.71          & 32.92          & 0.66          & {\ul 31.85}    & 0.58                & 6                         \\
                            & SSPSR  & 33.32          & 0.71          & 32.38          & 0.66          & 30.63          & 0.55                & 6                         \\
                            & MCNet  & {\ul 34.55}    & {\ul 0.74}    & {\ul 33.55}    & {\ul 0.69}    & 31.76          & {\ul 0.59}          & 22                        \\
                            & RFSR   & 33.70          & 0.72          & 32.73          & 0.66          & 31.06          & 0.56                & 15                        \\
                            & PDENet & 29.79          & 0.64          & 28.78          & 0.57          & 29.28          & 0.55                & 16                        \\
                            & Ours   & \textbf{36.67} & \textbf{0.81} & \textbf{34.68} & \textbf{0.74} & \textbf{32.20} & \textbf{0.60}       & 22                        \\ \hline
\multirow{8}{*}{Houston}    & DIP2d  & 27.67          & 0.61          & 27.61          & 0.61          & 27.15          & 0.60                & 178                       \\
                            & DIP3d  & 27.72          & 0.61          & 27.67          & 0.61          & 27.29          & 0.60                & 4841                      \\
                            & SFCSR  & 29.90          & 0.69          & 29.00          & 0.65          & 27.35          & 0.60                & 5                         \\
                            & SSPSR  & 29.99          & 0.72          & 29.12          & 0.67          & 27.49          & 0.61                & 5                         \\
                            & MCNet  & {\ul 30.52}    & {\ul 0.72}    & {\ul 29.60}    & {\ul 0.68}    & {\ul 28.03}    & {\ul 0.62}          & 16                        \\
                            & RFSR   & 30.27          & 0.71          & 29.22          & 0.66          & 27.64          & 0.62                & 10                        \\
                            & PDENet & 28.83          & 0.60          & 28.06          & 0.56          & 27.71          & 0.58                & 11                        \\
                            & Ours   & \textbf{31.83} & \textbf{0.78} & \textbf{30.68} & \textbf{0.72} & \textbf{29.10} & \textbf{0.65}       & 28                        \\ \hline
\multirow{8}{*}{Salinas}    & DIP2d  & {\ul 36.29}    & 0.88          & 34.35          & 0.84          & 30.72          & 0.75                & 85                        \\
                            & DIP3d  & 35.24          & 0.86          & 34.60          & 0.85          & 32.34          & 0.82                & 1866                      \\
                            & SFCSR  & 35.63          & 0.87          & 34.42          & 0.86          & 32.15          & 0.82                & 4                         \\
                            & SSPSR  & 33.71          & 0.86          & 32.40          & 0.84          & 30.08          & 0.79                & 5                         \\
                            & MCNet  & 36.01          & {\ul 0.89}    & {\ul 35.13}    & {\ul 0.87}    & {\ul 32.47}    & \textbf{0.82}       & 10                        \\
                            & RFSR   & 35.26          & 0.87          & 34.02          & 0.85          & 31.59          & 0.81                & 12                        \\
                            & PDENet & 31.45          & 0.70          & 30.52          & 0.67          & 31.14          & 0.75                & 8                         \\
                            & Ours   & \textbf{39.44} & \textbf{0.91} & \textbf{37.53} & \textbf{0.88} & \textbf{34.48} & {\ul \textbf{0.82}} & 14                        \\ \hline
\end{tabular}
}
\end{table}

\begin{table}[h]
\caption{The average quantitative results for noisy HSI inpainting. The \textbf{best} and \underline{second-best} values are highlighted.}
\label{table:inpainting}
\renewcommand\arraystretch{1.2}
\resizebox{\linewidth}{!}{
\begin{tabular}{cc|ccccccc}
\hline
\multicolumn{2}{c|}{Masking   Rate} & \multicolumn{2}{c}{0.7}        & \multicolumn{2}{c}{0.8}        & \multicolumn{2}{c}{0.9}        & \multirow{2}{*}{Time (s)} \\ \cline{1-8}
Dataset                    & Method & PSNR           & SSIM          & PSNR           & SSIM          & PSNR           & SSIM          &                           \\ \hline
\multirow{8}{*}{WDC mall}  & TRPCA  & 22.29          & 0.23          & 22.72          & 0.22          & 23.30          & 0.22          & 346                       \\
                           & TRLRF  & 22.03          & 0.27          & 20.61          & 0.19          & 11.79          & 0.03          & 1000                      \\
                           & S2NTNN & 35.64          & 0.80          & 33.70          & 0.73          & 31.03          & 0.62          & 32                        \\
                           & HLRTF  & 37.03          & 0.85          & 36.07          & 0.82          & {\ul 34.08}    & {\ul 0.72}    & 160                       \\
                           & DIP2d  & 31.55          & 0.57          & 31.54          & 0.57          & 30.71          & 0.56          & 50                        \\
                           & DIP3d  & 31.47          & 0.57          & 31.90          & 0.57          & 31.69          & 0.57          & 12876                     \\
                           & DDS2M  & \textbf{38.32} & \textbf{0.91} & {\ul 36.18}    & \textbf{0.85} & 32.35          & 0.71          & 4920                      \\
                           & Ours   & {\ul 37.90}    & {\ul 0.87}    & \textbf{36.82} & {\ul 0.83}    & \textbf{35.08} & \textbf{0.76} & 17                        \\ \hline
\multirow{8}{*}{Houston}   & TRPCA  & 22.13          & 0.21          & 22.46          & 0.21          & 22.84          & 0.22          & 175                       \\
                           & TRLRF  & 21.16          & 0.20          & 19.11          & 0.13          & 13.32          & 0.04          & 768                       \\
                           & S2NTNN & 31.01          & 0.76          & 28.08          & 0.67          & 24.48          & 0.46          & 16                        \\
                           & HLRTF  & 30.82          & 0.77          & 29.92          & 0.72          & 28.53          & 0.65          & 57                        \\
                           & DIP2d  & 26.45          & 0.59          & 26.35          & 0.58          & 26.19          & 0.58          & 48                        \\
                           & DIP3d  & 26.51          & 0.59          & 26.28          & 0.58          & 26.12          & 0.58          & 3037                      \\
                           & DDS2M  & {\ul 31.91}    & {\ul 0.81}    & {\ul 30.05}    & {\ul 0.73}    & {\ul 29.13}    & {\ul 0.67}    & 2638                      \\
                           & Ours   & \textbf{32.42} & \textbf{0.82} & \textbf{31.19} & \textbf{0.78} & \textbf{29.67} & \textbf{0.72} & 20                        \\ \hline
\multirow{8}{*}{Salinas}   & TRPCA  & 22.67          & 0.18          & 23.30          & 0.20          & 23.96          & 0.23          & 79                        \\
                           & TRLRF  & 22.75          & 0.19          & 20.42          & 0.13          & 13.09          & 0.03          & 324                       \\
                           & S2NTNN & 37.74          & 0.88          & 34.14          & 0.75          & 26.79          & 0.61          & 11                        \\
                           & HLRTF  & 33.97          & 0.79          & 33.40          & 0.78          & 32.56          & 0.77          & 31                        \\
                           & DIP2d  & 34.71          & 0.86          & 33.65          & 0.84          & 32.15          & 0.82          & 41                        \\
                           & DIP3d  & 33.76          & 0.85          & 33.04          & 0.83          & 31.77          & 0.81          & 1274                      \\
                           & DDS2M  & {\ul 40.49}    & \textbf{0.94} & {\ul 38.23}    & {\ul 0.91}    & {\ul 36.80}    & {\ul 0.89}    & 1354                      \\
                           & Ours   & \textbf{40.70} & {\ul 0.94}    & \textbf{39.40} & \textbf{0.92} & \textbf{37.61} & \textbf{0.90} & 16                        \\ \hline
\end{tabular}
}
\end{table}

\subsection{Ablation Study}\label{subsec:ablation}
\paragraph{Reduced Image Estimation}

\begin{figure}
  \centering
  \includegraphics[width=0.8\linewidth]{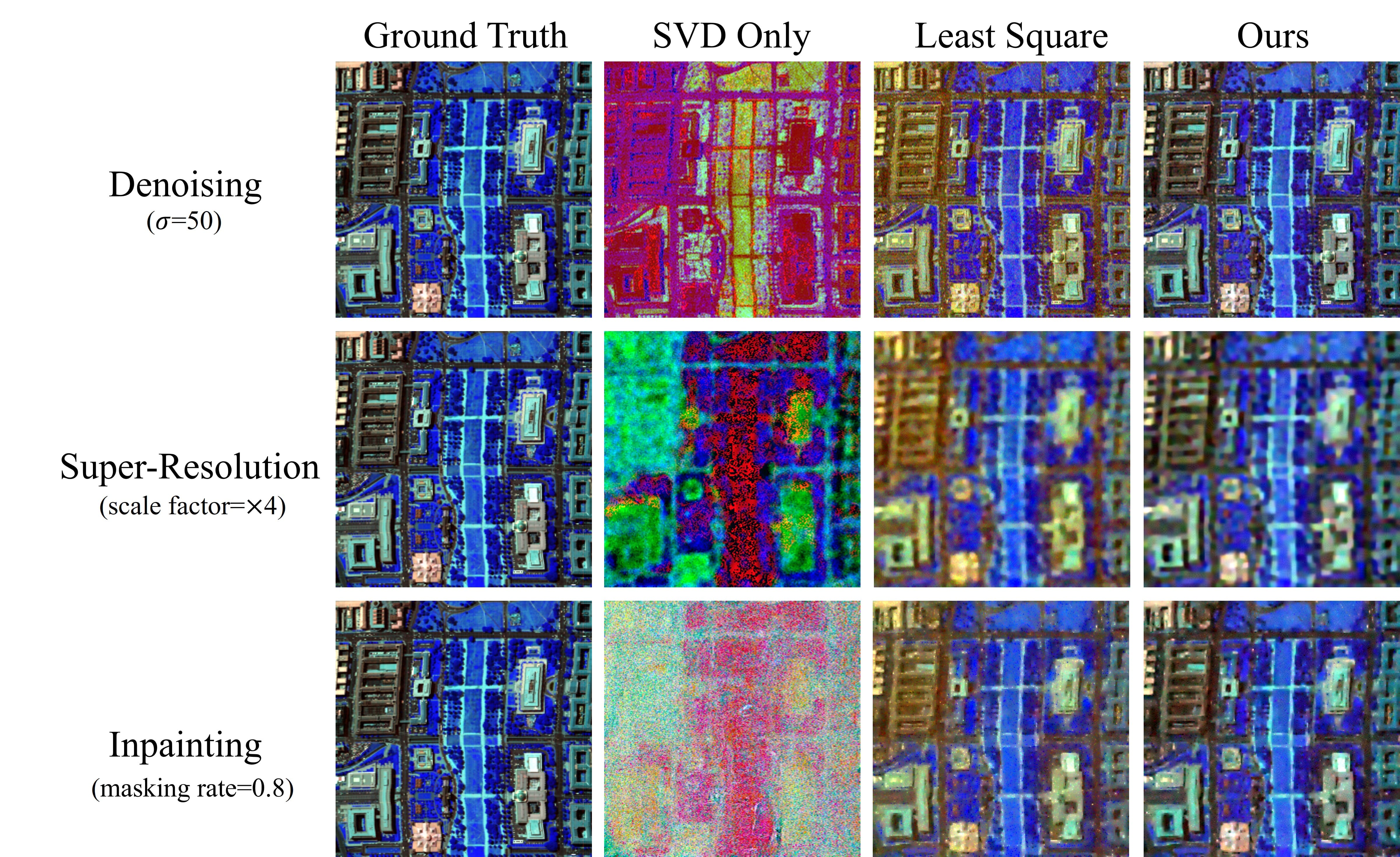}
  \caption{The reduced image $\mathcal{A}$ estimated by different methods. Ground Truth denotes the bands selected from the clean HSI.}\label{fig:reducedimage}
\end{figure}

In our work, an HSI is restored from a reduced image $\mathcal{A}$ which is supposed to be several bands, and a corresponding coefficient matrix $\mathbf{E}$. Nevertheless, there exists other alternative tensor decomposition methods to restore an HSI.
One of them is directly defining the $\mathbf{V}$ matrix obtained from SVD as the coefficient matrix $\mathbf{E}$, and employing the diffusion model to restore the "pseudo" image $\mathcal{A}=\mathbf{\text{fold}_3(US)}$.
Another solution \cite{rui2023unsupervised} is that the coefficient matrix $\mathbf{E}$ is estimated by solving a least square problem with several observed noisy bands, and then $\mathcal{A}$ is estimated from the diffusion model.
The definition of these decomposition strategies is inconsistent with the diffusion prior or is susceptible to noise, leading to poor restoration results. The HSI restoration results of different decomposition strategies on WDC dataset are illustrated in Table \ref{table:ablationA} and Fig.~\ref{fig:reducedimage}.
Our method demonstrates superior HSI restoration performance as the SVD operation effectively suppresses noise and the definition of the tensor $\mathcal{A}$ lies in the image field, which is consistent with the diffusion prior.

\begin{table}[]
\caption{The HSI restoration results of different tensor decomposition methods. "SVD Only" denotes that the tensor $\mathcal{A}$ is defined as the pseudo image of SVD and "Least Square" denotes that the matrix $\mathbf{E}$ is estimated by solving the least square problem.}
\label{table:ablationA}
\renewcommand\arraystretch{1.2}
\resizebox{\linewidth}{!}{
\begin{tabular}{lcccccc}
\hline
Task         & \multicolumn{2}{c}{Denoising}  & \multicolumn{2}{c}{Super-Resolution} & \multicolumn{2}{c}{Inpainting} \\ \hline
Methods      & PSNR           & SSIM          & PSNR              & SSIM             & PSNR           & SSIM          \\ \hline
SVD Only     & 28.88          & 0.55          & 27.86             & 0.43             & 28.16          & 0.44          \\
Least Square & 37.53          & 0.89          & 34.37             & 0.73             & 35.55          & 0.79          \\
Ours         & \textbf{40.73} & \textbf{0.94} & \textbf{34.79}    & \textbf{0.74}    & \textbf{37.02} & \textbf{0.85} \\ \hline
\end{tabular}
}
\end{table}

\begin{table}[h]
\caption{The HSI restoration results of different band selection indexes. $(\cdot, \cdot, \cdot)^*$ denotes the bands selected using our method.}
\label{table:ablationC}
\resizebox{\linewidth}{!}{
\begin{tabular}{llccc}
\hline
Task                              & Bands             & $|\det(\mathbf{V}_s)|$                & \tabincell{c}{The Maximum \\Value in $\mathbf{E}$} & PSNR  \\ \hline
\multirow{4}{*}{Denoising}        & (1, 48, 96)       & 0.000681                         & 7.91                              & 27.85 \\
                                  & (37, 84, 132)     & 0.001427                         & 5.86                              & 28.00 \\
                                  & (73, 120, 168)    & 0.000068                         & 11.57                             & 18.40 \\
                                  & $\mathbf{(21, 31, 60)^*}$  &  \textbf{0.011500}                         & \textbf{1.01}                              & \textbf{44.27} \\ \hline
\multirow{4}{*}{Super-Resolution} & (1, 48, 96)       & 0.000195                         & 24.86                             & 11.02 \\
                                  & (37, 84, 132)     & 0.000566                         & 10.28                             & 20.41 \\
                                  & (73, 120, 168)    & 0.000193                         & 12.58                             & 9.96  \\
                                  & $\mathbf{(22, 33, 58)^*}$ & \textbf{0.012835}                         & \textbf{1.00}                               & \textbf{39.68} \\ \hline
\multirow{4}{*}{Inpainting}       & (1, 48, 96)       & 0.001308                         & 3.11                              & 34.54 \\
                                  & (37, 84, 132)     & 0.000617                         & 16.07                             & 9.93  \\
                                  & (73, 120, 168)    & 0.000007                         & 47.34                             & 6.51  \\
                                  & \textbf{$\mathbf{(21, 34, 58)^*}$}  & \textbf{0.014779}                         & \textbf{1.00}                              & \textbf{41.02} \\ \hline
\end{tabular}
}
\end{table}
\paragraph{Coefficient Matrix Estimation}
As introduced in Sec.~\ref{subsec:matrix_estimation}, we employ RRQR to determine the band selection index $(i_1,i_2,...,i_K)$. The HSI restoration results on Salinas dataset of several indexes selected at equal intervals are provided in Table~\ref{table:ablationC} to prove the necessity of RRQR. As can be observed, due to the low-rank property of HSIs, the determinant of $\mathbf{V}_s$ could be extremely small, indicating a high similarity between the selected bands and thus resulting in large values in $\mathbf{E}$ and numerical instability. In contrast, the bands selected using our method encode more abundant information and provide desirable results.


\paragraph{Guidance Function}
The results of different guidance functions are provided in Table~\ref{table:ablationG}. As can be seen, the absence of any condition leads to the worst performance since the model becomes a pure generative model. In addition, the TV regularization adopted in our work contributes to the improvement of restoration performance, verifying the effectiveness of our proposed guidance function.
\begin{table}[h]
\caption{Ablation study of the proposed guidance function.}
\label{table:ablationG}
\resizebox{\linewidth}{!}{
\begin{tabular}{cc|cccccc}
\hline
\multirow{2}{*}{\tabincell{c}{Data Fidelity\\ Term}} & \multirow{2}{*}{\tabincell{c}{TV Regularization\\ Term}} & \multicolumn{2}{c}{Denoising}                         & \multicolumn{2}{c}{Super-Resolution}                & \multicolumn{2}{c}{Inpainting}                      \\
                                      &                                         & \multicolumn{1}{l}{PSNR} & \multicolumn{1}{l}{SSIM} & \multicolumn{1}{l}{PSNR} & \multicolumn{1}{l}{SSIM} & \multicolumn{1}{l}{PSNR} & \multicolumn{1}{l}{SSIM} \\ \hline
\ding{55}                                     &       \ding{55}                                 & 10.12                    & 0.07                     & 11.49                    & 0.16                     & 11.13                    & 0.08                     \\
\checkmark                                     &      \ding{55}                                  & 34.99                    & 0.87                     & 27.10                    & 0.45                     & 29.90                    & 0.67                     \\
\checkmark                                     & \checkmark                                       & \textbf{36.14}           & \textbf{0.90}            & \textbf{30.70}           & \textbf{0.72}            & \textbf{31.71}           & \textbf{0.80}            \\ \hline
\end{tabular}
}
\end{table}

\paragraph{Noise Schedule}
The results of different noise schedules are shown in Table~\ref{table:ablationN}. As can be observed, the restoration performance is improved with our exponential schedule and declines much slower when the number of sampling step decreases since there are more effective sampling steps as introduced in Sec.~\ref{subsec:diffusion_improvement}.

\begin{table}[h]
\caption{The HSI restoration results of different noise schedules.}
\label{table:ablationN}
\resizebox{\linewidth}{!}{
\begin{tabular}{c|l|cccccc}
\hline
\multirow{2}{*}{Task}             & Step $\times T$ & \multicolumn{2}{c}{20}         & \multicolumn{2}{c}{50}         & \multicolumn{2}{c}{100}        \\
                                  & Schedule        & PSNR           & SSIM          & PSNR           & SSIM          & PSNR           & SSIM          \\ \hline
\multirow{3}{*}{Denoising}        & Linear          & 34.34          & 0.86          & 35.18          & 0.88          & 35.70          & 0.89          \\
                                  & Cosine          & 34.61          & 0.86          & 35.46          & 0.88          & 35.86          & 0.89          \\
                                  & Exponential     & \textbf{36.01} & \textbf{0.90} & \textbf{36.10} & \textbf{0.90} & \textbf{36.14} & \textbf{0.90} \\ \hline
\multirow{3}{*}{Super-Resolution} & Linear          & 29.47          & 0.66          & 30.01          & 0.68          & 30.39          & 0.70          \\
                                  & Cosine          & 30.25          & 0.70          & 30.45          & 0.70          & 30.56          & 0.71          \\
                                  & Exponential     & \textbf{30.68} & \textbf{0.72} & \textbf{30.69} & \textbf{0.72} & \textbf{30.70} & \textbf{0.72} \\ \hline
\multirow{3}{*}{Inpainting}       & Linear          & 29.96          & 0.70          & 30.86          & 0.74          & 31.44          & 0.77          \\
                                  & Cosine          & 30.17          & 0.71          & 31.07          & 0.76          & 31.46          & 0.77          \\
                                  & Exponential     & \textbf{31.19} & \textbf{0.78} & \textbf{31.65} & \textbf{0.79} & \textbf{31.71} & \textbf{0.80} \\ \hline
\end{tabular}
}
\end{table}

%% file: sec/6_conclusion.tex
\section{Conclusion}
\label{sec:conclusion}
In this paper, we propose an unsupervised HSI restoration approach by employing a pre-trained diffusion model. Utilizing the low-rank property of HSIs, the clean image is restored by the product of a reduced image and a coefficient matrix. Specifically, the coefficient matrix can be pre-estimated from the observed image utilizing Singular Value Decomposition (SVD) and Rank-Revealing QR (RRQR), and the reduced image can be estimated from the pre-trained diffusion model with a newly designed guidance function. Additionally, the proposed exponential noise schedule is proved to be more reasonable for our conditional diffusion model and can significantly accelerate the sampling process. Our proposed method is a universal HSI restoration framework and can obtain better performance against other state-of-the-art methods on various HSI restoration tasks.

%% file: sec/X_supplementary.tex
\clearpage

\setcounter{page}{1}
\maketitlesupplementary
\setcounter{section}{0}
\setcounter{subsection}{0}

\section{Band Index Selection Using RRQR}
\label{sec:supplementary_RRQR}
\subsection{Notation}
The QR factorization of matrix $\mathbf{X}\in\mathbb{R}^{m\times n}$ is defined as
\begin{equation}\label{eq:qr}
\setlength{\arraycolsep}{1.5pt}
\mathbf{X} = \mathbf{QR} \equiv \mathbf{Q}\left[\begin{array}{cc} \mathbf{R}_{11} & \mathbf{R}_{12} \\ \mathbf{0} & \mathbf{R}_{22}\end{array}\right],
\end{equation}
where $\mathbf{Q} \in \mathbb{R}^{m\times m}$ is an orthogonal matrix and $\mathbf{R} \in \mathbb{R}^{m\times n}$ is an upper triangular matrix. Then $\mathcal{R}_k(\mathbf{X})$ is defined as
\begin{equation}\label{eq:r}
\setlength{\arraycolsep}{1.5pt}
\mathcal{R}_k(\mathbf{X}) = \left[\begin{array}{cc} \mathbf{R}_{11} & \mathbf{R}_{12} \\ \mathbf{0} & \mathbf{R}_{22}\end{array}\right].
\end{equation}
For $\mathbf{R}_{11}$, $1/\omega_i(\mathbf{R}_{11})$ denotes the 2-norm of the $i$th row of $\mathbf{R}_{11}^{-1}$.
For $\mathbf{R}_{22}$, $1/\gamma_i(\mathbf{R}_{22})$ denotes the 2-norm of the $j$th column of $\mathbf{R}_{22}$.
Then $\rho(\mathbf{R},k)$ is defined as
\begin{small}
\begin{equation}\label{eq:rho}
\rho(\mathbf{R},k)=\max\limits_{1\leq i\leq k \atop 1\leq j\leq n-k}\sqrt{\left(\mathbf{R}_{11}^{-1}\mathbf{R}_{12}\right)_{i,j}^2+\left(\gamma_j(\mathbf{R}_{22})/\omega_i(\mathbf{R}_{11})\right)^2}.
\end{equation}
\end{small}
$\mathbf{\Pi}_{i,j}$ denotes the permutation that interchanges the $i$th and $j$th columns of a matrix.

\subsection{RRQR}
In this section, we provide a more detailed description of the RRQR algorithm process~\cite{gu1996efficient} employed in our work. As introduced in Sect.~\ref{subsec:matrix_estimation}, the RRQR is used to determine the band index $(i_1,i_2,...,i_K)$ so that $|\det(\mathbf{V}_s)|$ is prevented from being zero and each band in the reduced image $\mathcal{A}$ is able to encode different image information. Given a matrix $\mathbf{M}\in\mathbb{R}^{m\times n}$ with $m \geq n$, the QR factorization of $\mathbf{M}$ with its columns permuted can be formulated as
\begin{equation}\label{eq:rrqr}
\setlength{\arraycolsep}{1.5pt}
\mathbf{M\Pi} = \mathbf{QR} \equiv \mathbf{Q}\left[\begin{array}{cc} \mathbf{R}_{11} & \mathbf{R}_{12} \\ \mathbf{0} & \mathbf{R}_{22}\end{array}\right],
\end{equation}
where $\mathbf{Q} \in \mathbb{R}^{m\times m}$ is an orthogonal matrix, $\mathbf{R} \in \mathbb{R}^{m\times n}$ is an upper triangular matrix and $\mathbf{\Pi}$ is a permutation matrix.

The RRQR aims to choose $\mathbf{\Pi}$ such that $\sigma_{\min}(\mathbf{R_{11}})$ is sufficiently large and $\sigma_{\max}(\mathbf{R_{22}})$ is sufficiently small, where $\sigma(\cdot)$ denotes the singular values. The RRQR factorization algorithm works by interchanging any pair of columns that sufficiently increases $|\det(\mathbf{R_{11}})|$, resulting in a large $|\det(\mathbf{R_{11}})|$. The details of the RRQR process are illustrated in Algorithm~\ref{alg:algorithm2}, where
$k$ and $f \geq 1$ are hyperparameters. It can be proven that $|\det(\mathbf{R_{11}})|$ increases strictly with every interchange and the algorithm is completed within a finite number of permutations. Readers could refer to~\cite{gu1996efficient} for more detailed proof.
\begin{algorithm}[h]
	\caption{RRQR Algorithm}
	\label{alg:algorithm2}
    \SetKw{True}{True}
    \KwIn{$\mathbf{M}, k, f$}
    \KwOut{Permutation matrix $\mathbf{\Pi}$}
	\BlankLine

    Compute $\mathbf{R}:=\mathcal{R}_k(M)$ and $\mathbf{\Pi}=\mathbf{I}$ \\
    \While{\True}{
    \textbf{step 1:} Compute $\rho(\mathbf{R},k)$ \\
    \textbf{step 2:} \lIf{$\rho(\mathbf{R},k) \leq f$}{\textbf{Break}}
    \textbf{step 3:} Find $i$ and $j$ such that $\rho(\mathbf{R},k)>f$ \\
    \textbf{step 4:} Compute $\mathbf{R}:=\mathcal{R}_k(\mathbf{R\Pi}_{i,j+k})$ and $\mathbf{\Pi}:=\mathbf{\Pi}\mathbf{\Pi}_{i,j+k}$
    }

    \Return{$\mathbf{\Pi}$}
\end{algorithm}
We replace $\mathbf{M}$ in Eq.(\ref{eq:rrqr}) with $\mathbf{V}^\mathrm{T}$ and we have
\begin{equation}\label{eq:rrqrV}
\setlength{\arraycolsep}{1.5pt}
\mathbf{V^\mathrm{T}\mathbf{\Pi}_v} = \left[\begin{array}{cc}\mathbf{Q}_v\mathbf{R}_{v1}&\mathbf{Q}_v\mathbf{R}_{v2}\end{array}\right]
\end{equation}
We define $\mathbf{V}_s^\mathrm{T}$ as $\mathbf{Q}_v\mathbf{R}_{v1}$ and then we could readily obtain $|\det(\mathbf{V}_s)|=|\det(\mathbf{R}_{v1})|$. Therefore, by solving the RRQR factorization problem of $\mathbf{V}^\mathrm{T}$ utilizing the algorithm proposed in~\cite{gu1996efficient}, the permutation matrix $\mathbf{\Pi}_v$ is obtained and $|\det(\mathbf{V}_s)|$ is maximized. The indices corresponding to the first $K$ columns of the permuted $\mathbf{V}^\mathrm{T}$ are defined as the band selection index.

\section{Diffusion Model}
In HIR-Diff, we employ a pre-trained diffusion model proposed in~\cite{gedara2022remote} to generate the reduced image $\mathcal{A}$. The diffusion model is a U-Net proposed in~\cite{saharia2022image} and is trained on an amount of 3-channel remote sensing images without human supervision. Since the network requires the input image to be 3-channel, the rank value $K$ in our work is set as 3 so that the reduced image $\mathcal{A}$ with 3 channels can be denoised with the pre-trained network. Although the rank value is small, we found that it is sufficient to restore the image details and helps to keep noise out of the estimated coefficient matrix $E$ and the restored image, since the matrix $\mathbf{V}$ obtained from the SVD of the observed image as introduced in Sec.~\ref{subsec:matrix_estimation} is cleaner and the low-rank property enables noise reduction of the restored image.

\begin{figure*}[h]
  \centering
  \includegraphics[width=\linewidth]{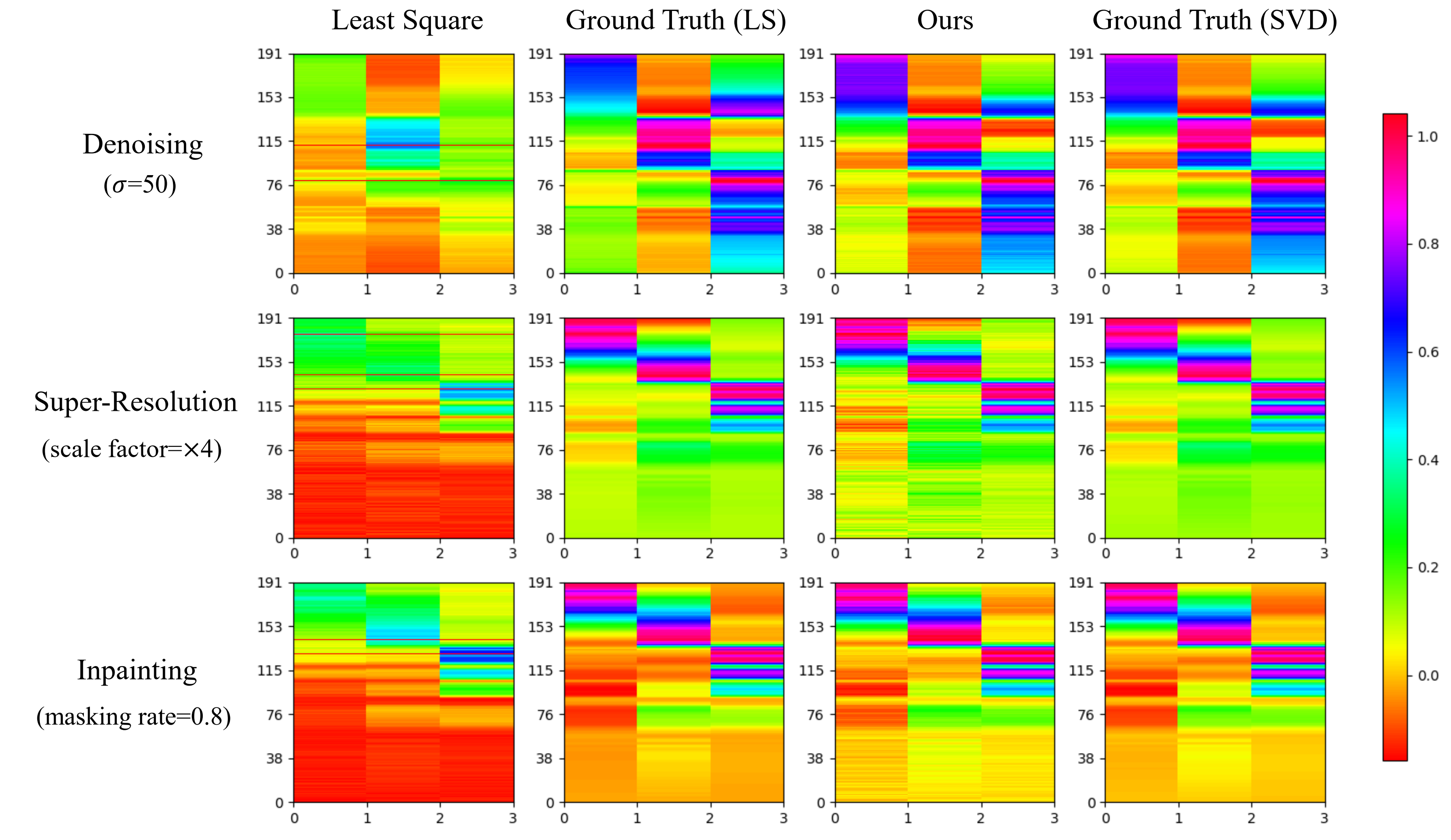}
  \caption{The visualization results of the estimated $\mathbf{E}$. \textbf{Least Square} and \textbf{Ground Truth (LS)} denote the coefficient matrix $\mathbf{E}$ estimated by employing the least square method with the observed image and the clean image, respectively. \textbf{Ours} and \textbf{Ground Truth (SVD)} denote the coefficient matrix $\mathbf{E}$ estimated using SVD and RRQR proposed in our work with the observed image and the clean image, respectively.
  }\label{fig:E}
\end{figure*}

\section{Coefficient Matrix Estimation}
\label{sec:supplementary_E}
In our work, we employ SVD and RRQR to estimate the coefficient matrix $\mathbf{E}$ as introduced in Sec.~\ref{subsec:matrix_estimation}. The visualization results of the estimated $\mathbf{E}$ for the WDC dataset are demonstrated in Fig.~\ref{fig:E}. The results of the matrix $\mathbf{E}$ estimated using the least square method proposed in~\cite{rui2023unsupervised} is also provided for comparison. Specifically, the least-squares method directly selects several bands from the observed image $\mathcal{Y}$ as the reduced image $\mathcal{A}$, and then estimates the coefficient matrix $E$ by solving the least-squares problem.
\begin{equation}\label{eq:ls}
\arg\min\limits_\mathbf{E}\left|\left|\mathcal{Y}-\mathcal{A}\times_3E\right|\right|_F^2.
\end{equation}
Since there is a lot of noise in the reduced image $\mathcal{A}$ as the observed image suffers from various degradation, the estimated $\mathbf{E}$ is unreliable, resulting in undesirable HSI restoration performance. On the contrary, our estimated coefficient matrix $\mathbf{E}$ is robust to noise and exhibits a high degree of similarity to ground truth, verifying the effectiveness of our estimation method.
